\documentclass{article}
\usepackage{graphicx} 
\usepackage{times}
\usepackage{subfigure} 
\usepackage[accepted]{icml2014} 
\usepackage{natbib}
\usepackage[all]{xy}
\begin{document}
\renewcommand{\baselinestretch}{1.1}
\newcommand{\jj}{{\bf j}}
\newcommand{\uu}{{\bf u}}
\newcommand{{\vv}}{{\bf v}}
\newcommand{\bga}{{\bf w}}
\newcommand{\rb}{\rangle}
\newcommand{\lb}{\langle}
\newcommand{\om}{\omega}
\newcommand{\Real} {{\rm Real}}

\twocolumn[
\icmltitle{Generic Deep Networks with Wavelet Scattering}

\icmlauthor{Edouard Oyallon}{edouard.oyallon@ens.fr}
\icmladdress{Ecole Normale Sup\'erieure, 45 rue d'Ulm, 75005 Paris, France}
\icmlauthor{St\'ephane Mallat}{stephane.mallat@ens.fr}
\icmladdress{Ecole Normale Sup\'erieure, 45 rue d'Ulm, 75005 Paris, France}
\icmlauthor{Laurent Sifre}{laurent.sifre@polytechnique.edu}
\icmladdress{Ecole Normale Sup\'erieure, 45 rue d'Ulm, 75005 Paris, France}

\icmlkeywords{Convolution Networks, Invariants, Wavelets, Scattering}

\vskip 0.3in
]

\begin{abstract} 
We introduce a two-layer wavelet scattering network, 
for object classification. 
This scattering transform computes a spatial wavelet
transform on the first layer and a new 
joint wavelet transform along spatial, angular
and scale variables in the second layer. 
Numerical experiments demonstrate that this two layer convolution network, 
which involves no learning and no max pooling, 
performs efficiently on complex image data sets such as
CalTech, with structural objects variability and clutter.
It opens the possibility to simplify deep neural network learning by
initializing the first layers with wavelet filters.
\end{abstract}

\section{Introduction}

Supervised training of deep convolution networks \cite{lecun1998gradient} 
is clearly highly effective for image classification, as shown by results on ImageNet \cite{krizhevsky2012imagenet}.
The first layers of the
networks trained on ImageNet also perform very well to classify images in
very different databases, which indicates that these layers capture generic 
image information \cite{zeiler2013visualizing,girshick2013rich,donahue2013decaf}.
This paper shows that such generic properties can be captured by
a scattering transform, which has the capability to build invariants to 
affine transformations.
Scattering transforms compute hierarchical invariants along groups of transformations by cascading wavelet convolutions and modulus non-linearities, along the group variables \cite{mallat2012group}.

Invariant scattering transforms to translations \cite{joanandmallat2013} and rotation-translations \cite{sifre2012combined} have previously been applied to digit recognition and texture discrimination. The main source of variability
of these images are due to deformations and stationary stochastic variability.
This paper applies a scattering transform to CalTech-101 and CalTech-256 
datasets, which
include much more complex structural variability of objects and clutter,
with good classification results. The scattering transform is adapted to the
important source of variability of these images, by applying wavelet 
transforms along spatial, rotation, and scaling variables, with separable
convolutions which is computationally efficient and by considering YUV 
color channels independently.

 It differs to most deep
network in two respects. No ad-hoc renormalization is added within the 
network and no max pooling is performed. All poolings are average pooling,
which guarantees the mathematical stability of the representation. It does not
affect the quality of results when applied to wavelets, even in cluttered
environments. This study concentrates on two layers because
adding a third layer of wavelet coefficients did not reduce
classification errors. Beyond the first two layers, which take care of 
translation, rotation and scaling variability, 
 seems necessary to learn the filters involved in 
the third and next layers to improve classification performances.

\section{Scattering along Translations, Rotations and Scales}

\begin{figure}
\vskip 0.2in
\begin{center}

$ \xymatrix@C-=0.5cm@R-=0.5cm{
    x \ar[r] & \framebox[21px][c]{\small{$|W_1|$}} \ar[r]  \ar[d] & U_1 x   \ar[r]&  \framebox[21px][c]{\small{$|W_2|$}}\ar[r] \ar[d]&U_2x \ar[r]&  S_2x   \\
  & S_0x&&S_1x&&
  }$

\caption{A scattering representation is computed by successively computing the
modulus of wavelet coefficients with $|W_1|,|W_2|$, followed by an average pooling $\phi_J$.}
\label{fig1}
\end{center}
\vskip -0.2in
\end{figure}
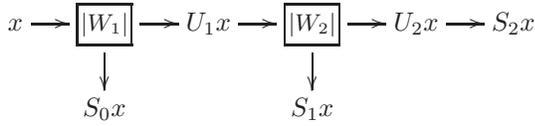

A two-layer 
scattering transform is computed by cascading wavelet transforms and modulus
non-linearities. The first wavelet transform $W_1$ filters the image $x$ with a low-pass filter
and complex wavelets which are scaled and rotated. The low-pass filter outputs an
averaged image $S_0 x$ and 
the modulus of each complex coefficients defines the first scattering layer $U_1 x$. 
A second wavelet transform $W_2$ applied to $U_1 x$ computes an average
$S_1 x$ and the next layer $U_2x$.
A final averaging 
computes second order scattering coefficients $S_2 x$, as illustrated in Figure \ref{fig1}. Higher order scattering coefficients are not computed.

The first wavelet transform is defined from a mother wavelet $\psi^1(u)$,
which is a complex Morlet function \cite{joanandmallat2013}
well localized in the image plane and a gaussian lowpass filter $\phi_J$.
The wavelet is scaled 
by $2^{j_1}$, where $j_1$ is an integer or half-integer, 
and rotated by $\theta_1 = 2 k \pi / K_1$ for $0 \leq k < K_1$:
\[
\psi^1_{\bga_1} (u) = 2^{-2j_1}\, \psi^1(2^{-2j_1} r_{\theta_1} u)
\]
where $\bga_1 = (\theta_1,j_1)$. The averaging filter is $\phi_J(u)=2^{-2J}\phi(2^{-J}u)$.

This wavelet transform first computes the average $S_0x=x\star\phi_J$ 
of $x(u)$ and we compute
the  modulus of the complex wavelet coefficients:
\begin{eqnarray*}
U_1 x(u,\bga_1) &=&  |x \star \psi^1_{\bga_1 }(u)| \\
&=&
 \Big|\sum_v x(u) \psi^1_{\bga_1} (u-v) \Big|~.
\end{eqnarray*}
The spatial variable $u$ is subsampled by $2^{j_1-1}$. 
We write $\uu_1 = (u,\bga_1)$ 
the aggregated variable which
indexes these first layer coefficients.

The next layer is computed with a second wavelet transform which convolves
$U_1 x(\uu_1)$ 
with separable wavelets along the spatial, rotation and
scale variables $\uu_1 = (u,\theta_1,j_1)$
\[
\psi^2_{\bga_2} (\uu_1) = \psi^1_{\theta_2,j_2} (u)\, \psi^b_{k_2} (\theta_1)\, \psi^c_{l_2} (j_1)~.
\]
The index $\bga_2 = (\theta_2,j_2,k_2,l_2)$ specifies the angle of rotation $\theta_2=2k\pi/K_2,0\leq k\leq K_2$, the scales $2^{j_2}$, $2^{k_2}$ and
$2^{l_2}$ of these wavelets. 
We choose this wavelet family so that it defines
a tight frame, and hence an invertible linear operator which preserves the
norm. The wavelet family in angle and scales also includes the necessary averaging filters. 
The next layer of coefficients are defined
for $\uu_1 = (u,\bga_1)$ and $\bga_2 = (\theta_2,j_2,k_2,l_2)$ by
\begin{eqnarray*}
U_2 x (\uu_1,\bga_2) &=& 
|U_1 x \star \psi^2_{\bga_2} (\uu_1)| \\
&=& 
\Big|\sum_{\vv_1} U_1 x(\vv_1) \,
\psi^2_{\bga_2} (\uu_1-\vv_1) \Big|~.
\end{eqnarray*}
First order order scattering coefficients are given by $S_1x=U_1x\star \phi_J$ and the second order scattering coefficients are computed with only a low-pass filtering $S_2x=U_2x\star \phi_J$ from the second layer. As opposed to almost all deep networks \cite{krizhevsky2012imagenet},
we found that no non-linear 
normalization was needed to obtain good classification
results with a scattering transform. As usual, at the classification stage,
all coefficients are standardized by setting their variance to $1$.

A locally invariant translation scattering representation is obtained 
by concatenating the 
scattering coefficients of different orders $S_0x,S_1x,S_2x$:
\[
S x  =\{S_0x,S_1x,S_2x\}.
\]
Each spatial averaging by $\phi_J$ is subsampled at intervals $2^{J-1}$.

\section{Numerical Classification Results}

The classification
performance of this double layer wavelet scattering representation is evaluated
on the Caltech databases. 
All images are first renormalized to a fixed size of $128$ by $128$ pixels
by a linear rescaling. Each YUV channel of each image is computed separately 
and their scattering coefficients are concatenated.
The first wavelet transform $W_1$ is computed 
with Morlet wavelets \cite{joanandmallat2013}, over
$5$ octaves $1 \leq 2^{j_1} \leq 2^5$ with $K_1 = 8$ angles $\theta_1$. 
The second wavelet transform $W_2$ is still 
computed with Morlet wavelets $\psi^1$
over a range of spatial scales $2^{j_1} \leq 2^{j_2} \leq 2^5$ and $K_2 = 8$ angles $\theta_2$.
We also use a Morlet wavelet $\psi^b$ over the angle variable, 
calculated over $3$ octaves $1 \leq 2^{l_2} \leq 2^3$. In this implementation, we did not use a wavelet along the scale variable $j_1$.

The final
scattering coefficients $S x$ are computed with a spatial pooling
at a scale $2^J = 32$, as opposed to the maxima selection used in most 
convolution networks. These coefficients are
renormalized by a standardization which subtracts their mean and sets their 
variance to $1$. The mean and variance are computed on the training
databases. Standardized scattering coefficients are then provided to 
a linear SVM classifier. 

\begin{table}[t]
\caption{Classification accuracies of convolution networks
on Caltech-101 and Caltech-256 using respectively 30  and 60 samples per class, depending upon the number of layers,
for a scattering transform and a network trained on ImageNet \cite{zeiler2013visualizing}.}
\label{sample-table}
\vskip 0.15in
\begin{center}
\begin{small}
\begin{sc}
\begin{tabular}{lcccr}
\hline
\abovespace\belowspace
Dataset & Layers & Calt.-101 & Calt.-256\\
\hline
\abovespace
Scattering & 1 & 51.2$\pm$0.8& 19.3$\pm$0.2\\
ImageNet CCN  &1 &44.8 $\pm$ 0.7&24.6$\pm$0.4 \\
Scattering & 2 & 68.8$\pm$ 0.5&34.6$\pm$0.2 \\
ImageNet CCN  & 2& 66.2$\pm$ 0.5&39.6$\pm$0.3 \\
ImageNet CCN  & 3& 72.3$\pm$ 0.4&46.0$\pm$0.3\\
ImageNet CCN  & 7& 85.5$\pm$ 0.4&72.6$\pm$0.2 \\
\hline
\end{tabular}
\end{sc}
\end{small}
\end{center}
\vskip -0.1in
\end{table}

Almost state of the art classification results 
are obtained on Caltech-101 and Caltech-256, 
with a ConvNet\cite{zeiler2013visualizing} pretrained on ImageNet.
Table \ref{sample-table} shows that 
with 7 layers it has an 85.5\% accuracy on Caltech-101 and 72.6\% accuracy on Caltech-256, using respectively 30 and 60 training images. The classification is performed with a linear SVM. In this work, we concentrate on the first two layers. With only two layers,
the ConvNet performances drop to $66.2\%$ on Caltech-101  and $39.6\%$ on Caltech-256, 
and progressively increase as the number of layers increases. 
A scattering transform 
has similar performances as a ConvNet when restricted to $1$ and $2$ layers,
as shown by Table \ref{sample-table}. It indicates that major sources of 
classification improvements over these first two layers can be obtained with wavelet
convolutions over spatial variables on the first layer, and joint spatial and
rotation variables on the second layer. Color improves by 1.5\% our results on Caltech-101 but it has not been tried yet on Caltech-256, whose results are given with gray-level images. More improvements can potentially be
obtained by adjusting the wavelet filtering along scale variables.  

Locality-constrained Linear Coding(LLC) \cite{wang2010locality} are examples different two layer
architectures, with a first layer computing SIFT feature vectors 
and a second layer which uses an unsupervised dictionary optimization.
This algorithm performs a max-pooling followed by an SVM.
LLC yields performances up to 73.4\% on Caltech-101 and 47.7\% on Caltech-256 
\cite{wang2010locality}. These results are better than the one obtained with fixed architectures such as the one presented in this paper, because the second layer performs an unsupervised optimization adapted to the data set.

In this work we observed that average-pooling can provide competitive and even better results that max-polling. According to some analysis and numerical experiments performed on sparse features \cite{boureau2010theoretical}, max-pooling should perform better than average-pooling. Caltech images are piecewise regular so we do have sparse wavelets coefficients. It however seems that using the modulus on complex wavelet coefficients does improves results obtained with average pooling compared to max-pooling. When using real wavelets and an absolute value non-linearity, average and max pooling achieve similar performances. The max-pooling was implemented with and without overlapping windows. Using squared windows of length $2^5,2^6,2^7$, with a max-pooling we obtained 60.5\% of accuracy on Caltech-101 without overlapping windows and 62.0\% with overlapping windows. This accuracy is well below the average pooling results presented in Table \ref{sample-table}. With real Haar wavelets, we observed that max-pooling and average pooling perform similarly with respectively 66.9\% and 67.0\% accuracy on Caltech-101. These difference behaviors with real and complex wavelets are still not well understood. 

\section{Conclusion}

We showed that a two layer scattering convolution network, which involves no learning, provides similar accuracy on the Caltech databases, 
as double layers neural network pretrained on ImageNet. 
This scattering transform linearizes
the variability relatively to translations and rotations and 
provides invariants to translations with an average pooling. 
It involves no inner renormalization but a standardisation of output
coefficients.

Many further improvements can still be brought to these first scattering layers, in 
particular by optimizing the scaling invariance. 
These preliminary experiments indicate that wavelet scattering transforms 
provide a good approach
to understand the first two layers of convolution networks for 
complex image classification, and an efficient initialization of these two layers.
\bibliography{myrefs}

\begin{thebibliography}{10}
\providecommand{\natexlab}[1]{#1}
\providecommand{\url}[1]{\texttt{#1}}
\expandafter\ifx\csname urlstyle\endcsname\relax
  \providecommand{\doi}[1]{doi: #1}\else
  \providecommand{\doi}{doi: \begingroup \urlstyle{rm}\Url}\fi

\bibitem[Boureau et~al.(2010)Boureau, Ponce, and LeCun]{boureau2010theoretical}
Boureau, Y-Lan, Ponce, Jean, and LeCun, Yann.
\newblock A theoretical analysis of feature pooling in visual recognition.
\newblock In \emph{Proceedings of the 27th International Conference on Machine
  Learning (ICML-10)}, pp.\  111--118, 2010.

\bibitem[Bruna \& Mallat(2013)Bruna and Mallat]{joanandmallat2013}
Bruna, J. and Mallat, S.
\newblock Invariant scattering convolution network.
\newblock \emph{IEEE Trans. on PAMI}, 35\penalty0 (8):\penalty0 1872--1886,
  2013.

\bibitem[Donahue et~al.(2013)Donahue, Jia, Vinyals, Hoffman, Zhang, Tzeng, and
  Darrell]{donahue2013decaf}
Donahue, J., Jia, Y., Vinyals, O., Hoffman, J., Zhang, N., Tzeng, E., and
  Darrell, T.
\newblock Decaf: A deep convolutional activation feature for generic visual
  recognition.
\newblock \emph{arXiv preprint arXiv:1310.1531}, 2013.

\bibitem[Girshick et~al.(2013)Girshick, Donahue, Darrell, and
  Malik]{girshick2013rich}
Girshick, R., Donahue, J., Darrell, T., and Malik, J.
\newblock Rich feature hierarchies for accurate object detection and semantic
  segmentation.
\newblock \emph{arXiv preprint arXiv:1311.2524}, 2013.

\bibitem[Krizhevsky et~al.(2012)Krizhevsky, Sutskever, and
  Hinton]{krizhevsky2012imagenet}
Krizhevsky, A., Sutskever, I., and Hinton, G.
\newblock Imagenet classification with deep convolutional neural networks.
\newblock In \emph{Advances in Neural Information Processing Systems 25}, pp.\
  1106--1114, 2012.

\bibitem[LeCun et~al.(1998)LeCun, Bottou, Bengio, and
  Haffner]{lecun1998gradient}
LeCun, Y., Bottou, L., Bengio, Y., and Haffner, P.
\newblock Gradient-based learning applied to document recognition.
\newblock \emph{Proceedings of the IEEE}, 86\penalty0 (11):\penalty0
  2278--2324, 1998.

\bibitem[Mallat(2012)]{mallat2012group}
Mallat, S.
\newblock Group invariant scattering.
\newblock \emph{Communications on Pure and Applied Mathematics}, 65\penalty0
  (10):\penalty0 1331--1398, 2012.

\bibitem[Sifre \& Mallat(2012)Sifre and Mallat]{sifre2012combined}
Sifre, L. and Mallat, S.
\newblock Combined scattering for rotation invariant texture analysis.
\newblock In \emph{European Symposium on Artificial Neural Networks}, 2012.

\bibitem[Wang et~al.(2010)Wang, Yang, Yu, Lv, Huang, and
  Gong]{wang2010locality}
Wang, Jinjun, Yang, Jianchao, Yu, Kai, Lv, Fengjun, Huang, Thomas, and Gong,
  Yihong.
\newblock Locality-constrained linear coding for image classification.
\newblock In \emph{Computer Vision and Pattern Recognition (CVPR), 2010 IEEE
  Conference on}, pp.\  3360--3367. IEEE, 2010.

\bibitem[Zeiler \& Fergus(2013)Zeiler and Fergus]{zeiler2013visualizing}
Zeiler, M.~D. and Fergus, R.
\newblock Visualizing and understanding convolutional neural networks.
\newblock \emph{arXiv preprint arXiv:1311.2901}, 2013.

\end{thebibliography}
\bibliographystyle{icml2014}

\end{document}